%% file: neurips_2025.tex
\documentclass{article}

\usepackage[preprint]{neurips_2025}

\usepackage[utf8]{inputenc} 
\usepackage[T1]{fontenc}    
\usepackage{hyperref}       
\usepackage{url}            
\usepackage{booktabs}       
\usepackage{amsfonts}       
\usepackage{nicefrac}       
\usepackage{microtype}      
\usepackage{xcolor}         

\usepackage{wrapfig} 
\usepackage{multirow}  
\usepackage{adjustbox}
\usepackage{subcaption}
\usepackage{caption}
\usepackage{amsmath}
\usepackage{pifont}
\title{SELFI: Selective Fusion of Identity for Generalizable Deepfake Detection}

\author{
  Younghun Kim\quad
  Minsuk Jang\quad
  Myung-Joon Kwon\quad
  Wonjun Lee\quad
  Changick Kim \\
  Korea Advanced Institute of Science and Technology (KAIST) \\
    Daejeon, South Korea \\
  \texttt{\{younghun1664, minsukjang, kwon19, dpenguin, changick\}@kaist.ac.kr}
}

\begin{document}

\maketitle

\begin{abstract}
Face identity provides a remarkably powerful signal for deepfake detection. Prior studies have shown that even when not explicitly modeled, deepfake classifiers tend to implicitly learn identity features during training. This has led to two conflicting viewpoints in the literature: some works attempt to completely suppress identity cues to mitigate bias, while others rely on them exclusively as a strong forensic signal. To reconcile these opposing stances, we conduct a detailed empirical analysis based on two central hypotheses: (1) whether face identity alone is inherently discriminative for detecting deepfakes, and (2) whether such identity features generalize poorly across manipulation methods.
Through extensive experimentation, we confirm that face identity is indeed a highly informative signal—but its utility is context-dependent. While some manipulation methods preserve identity-consistent artifacts, others distort identity cues in ways that can harm generalization. These findings suggest that identity features should not be suppressed or relied upon blindly. Instead, they should be explicitly modeled and adaptively controlled based on their per-sample relevance.
To this end, we propose \textbf{SELFI} (\textbf{SEL}ective \textbf{F}usion of \textbf{I}dentity), a generalizable deepfake detection framework that dynamically modulates identity usage. SELFI consists of: (1) a Forgery-Aware Identity Adapter (FAIA) that explicitly extracts face identity embeddings from a frozen face recognition model and projects them into a forgery-relevant space using auxiliary supervision, and (2) an Identity-Aware Fusion Module (IAFM) that selectively integrates identity and visual features via a relevance-guided fusion mechanism.
Extensive experiments on four benchmark datasets demonstrate that SELFI achieves strong generalization across manipulation methods and datasets, outperforming prior state-of-the-art methods by an average of 3.1\% frame-level AUC in cross-dataset evaluations. Notably, on the challenging DFDC benchmark, SELFI improves over the previous best by a significant 6\% margin, highlighting the effectiveness of adaptive identity control.
The code will be released upon acceptance of the paper.
\end{abstract}

\input{main_tex/Introduction}
\input{main_tex/RelatedWorks}
\input{main_tex/Preliminary}

\input{main_tex/Method}

\input{main_tex/Experiments}

\bibliographystyle{plain}
\bibliography{AT}

\newpage
\appendix
\input{main_tex/appendix}

\end{document}

%% file: main_tex/Introduction.tex
\begin{figure*}[t]
\centering
    \includegraphics[width=1\textwidth] {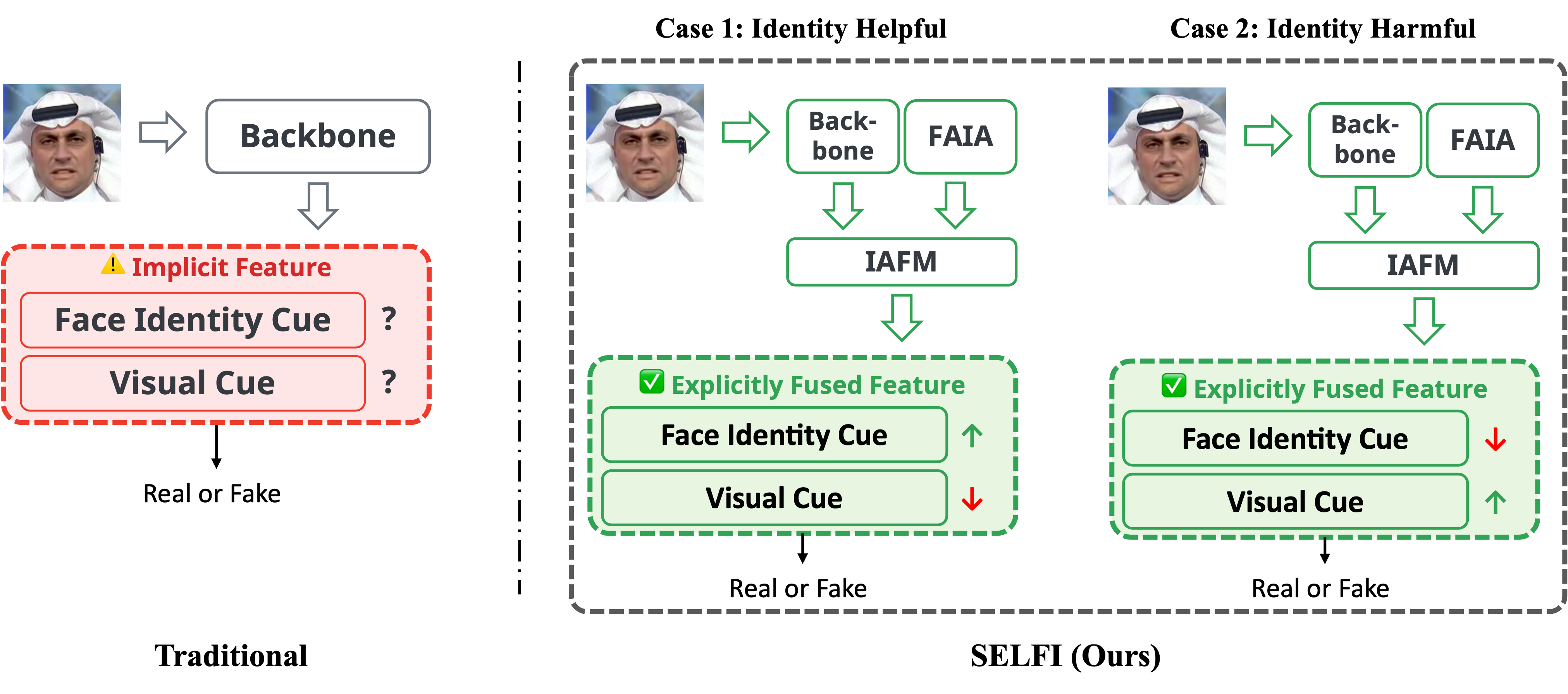}\\
    
    \caption{
\textbf{Overview of the proposed SELFI framework.}
Traditional deepfake detectors implicitly rely on identity cues without control, risking overfitting and poor generalization across manipulation types. In contrast, SELFI explicitly incorporates face identity features through a Forgery-Aware Identity Adapter (FAIA) and adaptively fuses them with visual features using an Identity-Aware Fusion Module (IAFM). This enables the model to leverage identity information when helpful and suppress it when harmful, resulting in more robust and generalizable deepfake detection.
    }
    \label{fig:figure1}
\end{figure*}

\section{Introduction}
\label{sec:intro}

Recent advances in deepfake generation~\cite{kowalski2018faceswap, thies2016face2face, li2019faceshifter, faceswapdevs2019deepfakes} have raised serious concerns about the authenticity of visual media, driving a surge in research on robust and generalizable detection methods~\cite{yan2024transcending, yan2023ucf, dong2022protecting, cheng2024can, cheng2024stacking}. While many existing detectors perform well on seen manipulation types, they often struggle to generalize to unseen forgeries—a critical limitation for real-world deployment~\cite{afchar2018mesonet, li2018exposing, yang2019exposing, qian2020thinking}. One possible explanation for this generalization gap lies in the treatment of face identity during training. Dong et al.~\cite{dong2023implicit} attribute poor generalization to implicit bias caused by face identity: because identity is such a strong discriminative signal, models tend to rely on it unconsciously, leading to overfitting to identity-specific patterns in the training data. In contrast, Huang et al.~\cite{huang2023implicit} argue that face identity itself is a powerful forensic cue, and that explicitly leveraging identity embeddings during training enhances robustness. These two opposing views—treating identity as a harmful bias versus a beneficial signal—highlight the need for a deeper understanding of how face identity influences detection performance across manipulation types.

To address this, we conduct an in-depth analysis guided by two core hypotheses. First, we examine whether face identity alone can meaningfully support deepfake detection. Specifically, we extract identity embeddings from a pretrained face recognition model (IResNet100~\cite{deng2019arcface}) that has never seen manipulated data, and train a lightweight classifier on top. Despite the absence of visual content and deepfake-specific supervision, we observe strong classification performance, confirming that face identity is inherently discriminative for this task. Second, we investigate whether such identity features generalize across manipulation methods, or if they encode method-specific patterns that hinder transferability. Using the FaceForensics++~\cite{rossler2019faceforensics++} dataset, we perform a cross-manipulation experiment where models trained on one type of forgery are tested on the others. This reveals three distinct identity behavior categories: (1) {transferable identity cues}, observed in manipulations like DeepFakes~\cite{faceswapdevs2019deepfakes} and FaceSwap~\cite{kowalski2018faceswap}, where identity information remains relatively intact and generalizes well; (2) {method-specific identity artifacts}, found in Face2Face~\cite{thies2016face2face}, where identity features are entangled with generation-specific artifacts and do not transfer; and (3) {ineffective identity cues}, as in NeuralTextures~\cite{thies2019deferred}, where identity features remain ambiguous, offering limited discriminative value for detection. These observations collectively indicate that the utility of identity features is highly context-dependent, and thus demand a more flexible, selective integration mechanism. In particular, while face identity is undeniably a powerful signal that should not be disregarded, allowing the model to implicitly absorb it without control can lead to generalization issues, especially when identity cues interact differently across manipulation methods.

These observations collectively indicate that the utility of identity features is highly context-dependent, and thus demand a more flexible, selective integration mechanism. In particular, while face identity is undeniably a powerful signal that should not be disregarded, allowing the model to implicitly absorb it without control can lead to generalization issues, especially when identity cues interact differently across manipulation methods. To explicitly address the challenges highlighted by this analysis regarding the context-dependent nature of identity cues and the necessity for selective integration, we design a solution that enables explicit and context-aware identity control. We propose SELFI (\textbf{SEL}ective \textbf{F}usion of \textbf{I}dentity), a novel framework that dynamically combines face identity features and backbone features to produce highly discriminative yet generalizable representations by adaptively leveraging identity based on its estimated relevance. SELFI consists of two core modules:
(1) the \textbf{F}orgery-\textbf{A}ware \textbf{I}dentity \textbf{A}dapter (FAIA), which projects identity embeddings extracted from a frozen face recognition model into a forgery-relevant representation space using a learnable transformation matrix. This projection is guided by an auxiliary supervision signal, the \textit{Forgery-Aware Guidance Loss}, which encourages the projected features to be discriminative for real-vs-fake classification even without visual cues, thereby making the identity information more pertinent to the detection task.
(2) the \textbf{I}dentity-\textbf{A}ware \textbf{F}usion \textbf{M}odule (IAFM), which adaptively integrates identity and visual features. IAFM includes a \textit{Relevance Predictor}, a lightweight neural network that estimates the importance of identity cues for each input, and a \textit{Soft Fusion Operator} that combines the two feature types via a weighted sum based on the predicted relevance score. This design allows SELFI to amplify identity information when it provides genuine forensic signals, and to downweight it when it risks introducing manipulation-specific bias, directly implementing the selective integration mechanism identified as necessary. The resulting fused feature representation jointly captures content and identity cues in a task-adaptive manner, enabling robust and generalizable deepfake detection. Fig.~\ref{fig:figure1} illustrates the overall comparison between traditional approaches and our SELFI framework under different identity relevance scenarios.

Our main contributions are summarized as follows:
\begin{itemize}
\item We conduct a quantitative analysis using a pretrained face recognition model, showing that identity features alone can distinguish real from fake content, and further identify three behavioral patterns across manipulation types: transferable,  method-specific identity cues, and ineffective.
\item Motivated by this observation, we propose SELFI, a framework that explicitly separates identity features from visual content and adaptively fuses them through two modules: FAIA, which projects identity embeddings into a forgery-discriminative space, and IAFM, which dynamically adjusts their contribution based on per-sample relevance.
\item Extensive experiments demonstrate that SELFI outperforms existing methods in both in-domain and cross-domain settings. Specifically, SELFI achieves an average improvement of {3.1\%} in frame-level AUC across four cross-dataset benchmarks (CDFv2, DFD, DFDC, DFDCP), and surpasses the previous best on {DFDC} by a significant {6.0\%} margin.

\end{itemize}

%% file: main_tex/RelatedWorks.tex
\section{Related Works}

\textbf{Deepfake Detection and Generalization.}
Early deepfake detection approaches primarily target visual artifacts introduced by manipulation techniques, such as blending boundaries~\cite{li2020face}, inconsistent textures~\cite{wang2023noise}, and abnormal frequency patterns~\cite{gu2022exploiting, liu2021spatial, luo2021generalizing, wang2023dynamic}. While these methods achieve promising performance on seen manipulation types, they often struggle to generalize to unseen forgeries or datasets due to their reliance on spurious and non-transferable cues. To enhance generalization, various strategies emerge. Frequency-domain methods aim to capture spectral inconsistencies between real and fake content~\cite{qian2020thinking, durall2020watch, luo2021generalizing, wang2023dynamic, li2024freqblender}, while spatiotemporal modeling approaches leverage temporal dynamics and cross-modal signals to detect subtle manipulation traces~\cite{wang2023altfreezing, xu2023tall, gu2022hierarchical}. Other studies employ latent space augmentation~\cite{yan2024transcending, choi2024exploiting} or self-blending techniques~\cite{shiohara2022detecting, li2020face} to diversify training data, and disentangle task-relevant features using representation separation~\cite{yan2023ucf, yang2023crossdf} or progressive regularization~\cite{cheng2024can}. Nevertheless, most of these methods focus predominantly on low-level visual cues, overlooking face identity—a semantically rich and manipulation-sensitive signal that remains underexplored despite its potential to offer deeper insights into forgery characteristics.

\textbf{Face Identity in Deepfake Detection.}
Several studies explore the role of facial identity in deepfake detection, revealing two contrasting perspectives. One line of research actively leverages face identity as a strong semantic cue to improve detection robustness. For instance, Dong et al.~\cite{dong2022protecting} propose ICT, an identity consistency transformer that measures coherence between inner and outer face regions, while Huang et al.~\cite{huang2023implicit} introduce IID, which defines implicit identity to detect face swapping by contrasting it with explicit identity embeddings. These methods demonstrate that explicitly modeling identity cues reveals forgery-specific inconsistencies. However, overreliance on face identity may cause models to overlook important visual cues, limiting their ability to capture manipulation artifacts beyond identity mismatches. In contrast, other works identify facial identity as a source of bias that harms generalization across datasets. Dong et al.~\cite{dong2023implicit} show in IIL that binary classifiers unintentionally learn identity boundaries, leading to implicit identity leakage and overfitting. Similarly, Kim et al.~\cite{kim2024friday} propose FRIDAY, which suppresses facial identity embeddings during training by encouraging the detector to diverge from a frozen face recognizer, thereby improving cross-domain robustness.

Motivated by these opposing findings, we aim to reconcile the contrasting views on face identity by conducting a systematic analysis grounded in two key hypotheses: (1) whether identity embeddings alone carry sufficient forensic information to detect manipulations, and (2) whether such features generalize reliably across different forgery methods. Our empirical findings reveal that face identity is indeed a highly discriminative signal, but its utility varies significantly depending on the manipulation context.




%% file: main_tex/Preliminary.tex
\begin{figure*}[t]
\centering
    \begin{minipage}[c]{.5\textwidth}
        \centering
        \includegraphics[width=\linewidth]{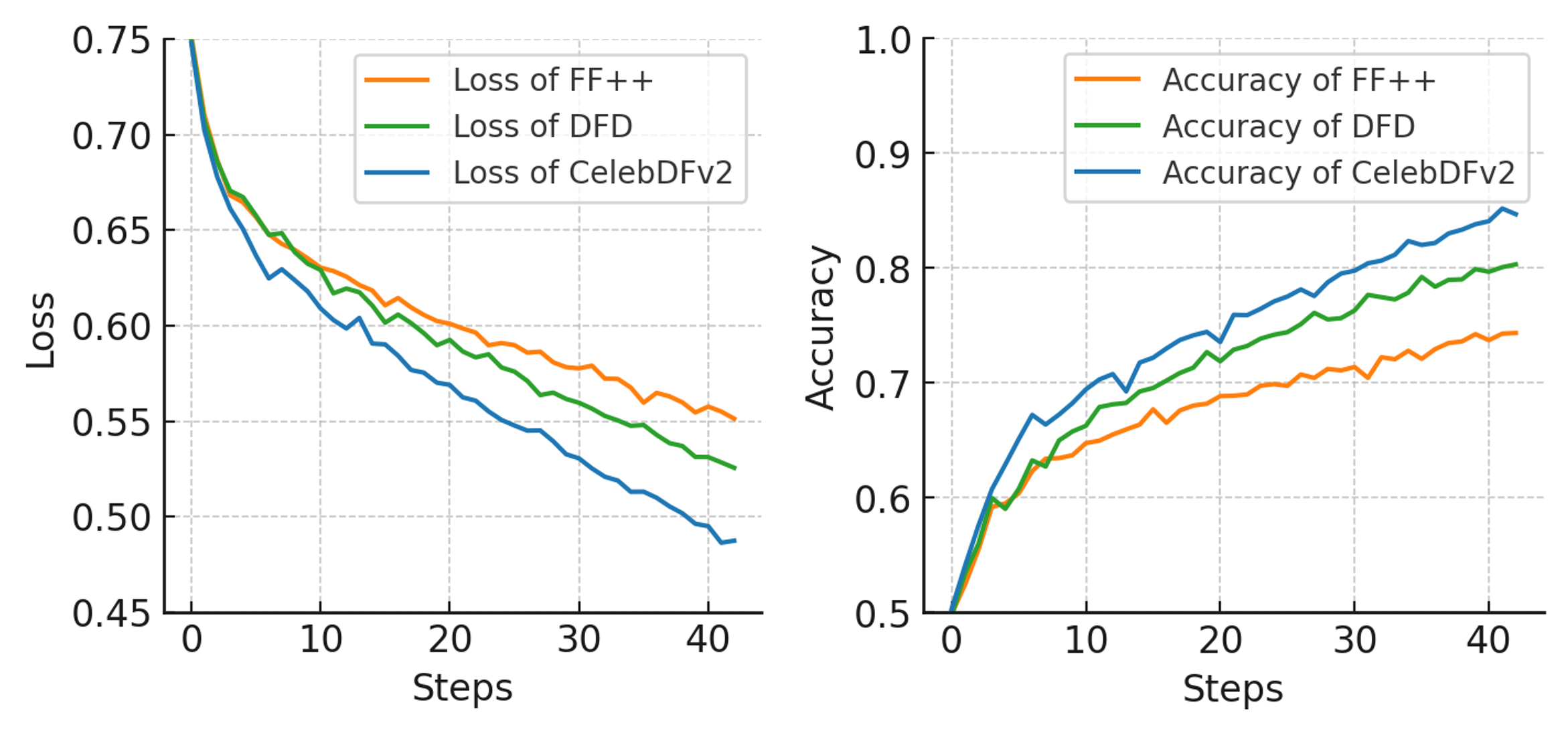}
        \subcaption{Training Loss and Accuracy}
    \end{minipage}%
    \hfill
    \begin{minipage}[c]{.5\textwidth}
        \centering
        \includegraphics[width=\linewidth]{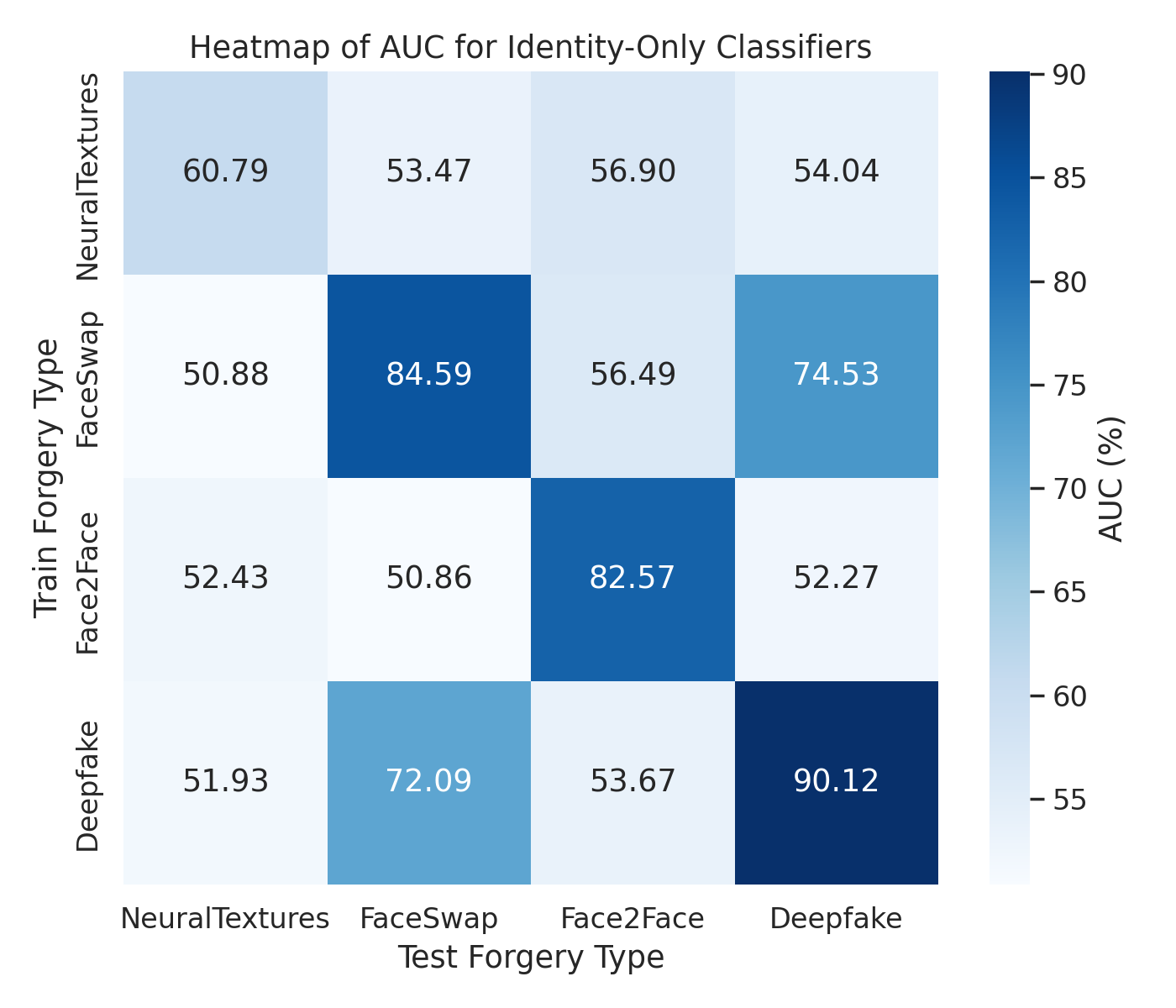}
        \subcaption{Cross-Manipulation AUC}
    \end{minipage}

    \caption{\textbf{Preliminary Analysis of Identity-Only Detection.} 
    (a) Training loss and accuracy curves for identity-only detection using a frozen face recognition model (IResNet100) on FF++, DFD, and CelebDFv2. The results show consistent decreases in loss and increases in accuracy, indicating that face identity alone provides strong discriminative power for real-vs-fake classification. 
    (b) Cross-manipulation AUC heatmap, where rows indicate training forgery types and columns indicate test types. The results reveal that identity features are transferable for some manipulations (DF and FS), but highly method-specific (F2F) or ineffective for others (NT), highlighting the need for explicit identity control.}

    \label{fig:preliminary}
\end{figure*}

\section{Preliminary Analysis}
\label{sec:preliminary}

\subsection{The Power of Face Identity in Deepfake Detection}
\label{subsec:hypothesis1}
\textbf{Hypothesis 1.} Face identity itself serves as a highly discriminative signal for deepfake detection, even without explicit modeling of forgery artifacts.

While face identity has long been considered an important cue for deepfake detection—albeit often implicitly—recent studies have expressed conflicting views on its role~\cite{dong2023implicit, huang2023implicit, kim2024friday, dong2022protecting}. Some suggest that identity is a reliable signal for detecting manipulations, while others argue that reliance on identity introduces undesirable bias that hinders generalization. Motivated by this controversy, we aim to explicitly evaluate whether face identity alone carries sufficient forensic information to support deepfake detection.

To this end, we formulate Hypothesis~1 and conduct an experiment using identity embeddings extracted from a pretrained face recognition model that has never been exposed to deepfake data. Specifically, we adopt IResNet100~\cite{deng2019arcface}, a widely used and well-established identity representation model, and freeze all of its parameters. We then train only a lightweight classifier head on top of the identity embeddings to assess whether face identity alone can support deepfake classification. Detailed experimental settings, architectural illustration, and analysis of the training outcomes beyond loss or accuracy curves are provided in Sec.~\ref{appendix:preliminary}.

We conduct training and evaluation on three widely-used deepfake detection benchmarks: FaceForensics++~\cite{rossler2019faceforensics++}, Celeb-DF v2~\cite{li2020celeb}, and DeepFakeDetection~\cite{GoogleAI2019}. As shown in Fig.~\ref{fig:preliminary} (a), the classification loss consistently decreases and the accuracy steadily increases across all datasets, even with the backbone completely frozen. These result provide strong evidence that face identity contains inherently discriminative signals for real-vs-fake classification, and can serve as a standalone foundation for deepfake detection models.
We next investigate whether the identity features captured in this setting remain robust across different manipulation types, or if they instead encode method-specific biases that compromise generalization.

\subsection{The Generalization Risk of Identity-Specific Features}
\label{subsec:hypothesis2}
\textbf{Hypothesis 2.} Different deepfake generation methods manipulate face identity in distinct ways. Therefore, over-reliance on identity cues that work well for specific methods may hinder generalization to unseen manipulations.

To test this hypothesis, we conduct a cross-manipulation experiment using the FaceForensics++ dataset~\cite{rossler2019faceforensics++}, which includes four representative manipulation types: DeepFakes~\cite{faceswapdevs2019deepfakes}, FaceSwap~\cite{kowalski2018faceswap}, Face2Face~\cite{thies2016face2face}, and NeuralTextures~\cite{thies2019deferred}. For each method, we train a classifier using identity embeddings extracted from a frozen IResNet100 model, and evaluate both in-domain and cross-domain performance. For example, a model trained on DeepFakes is tested not only on DeepFakes but also on the other three types.

The results, illustrated in Fig.~\ref{fig:preliminary} (b), reveal three distinct identity behavior patterns: (1) \textit{Transferable identity cues}, as seen in DeepFakes and FaceSwap, where identity information is well preserved and generalizes effectively across methods (e.g., 84.6\% $\rightarrow$ 74.5\%); (2) \textit{Method-specific identity artifacts}, as in Face2Face, where in-domain performance is high (82.6\%) but generalization is poor (e.g., 52.3\% on DeepFakes), indicating overfitting to generation-specific patterns; and (3) \textit{Ineffective identity cues}, as in NeuralTextures, which produces weak or distorted identity features, leading to both low in-domain (60.8\%) and cross-domain (e.g., 53.5\%) performance.

These findings demonstrate that the effectiveness of identity features is highly dependent on the manipulation method. Some techniques preserve identity in a way that supports generalization, while others entangle identity with artifacts or suppress it entirely. Consequently, uniformly relying on identity cues can harm generalization when such cues encode method-specific biases.

We argue that a more robust strategy is to explicitly separate identity from visual content and adaptively regulate its use based on context. This insight directly motivates the design of our proposed framework, \textbf{SELFI}, which selectively integrates identity features according to their per-sample relevance.

%% file: main_tex/Method.tex
\section{Proposed Method}
\begin{figure*}[t]
\centering
    \vspace*{-0.4cm}
    \includegraphics[width=0.98\textwidth]{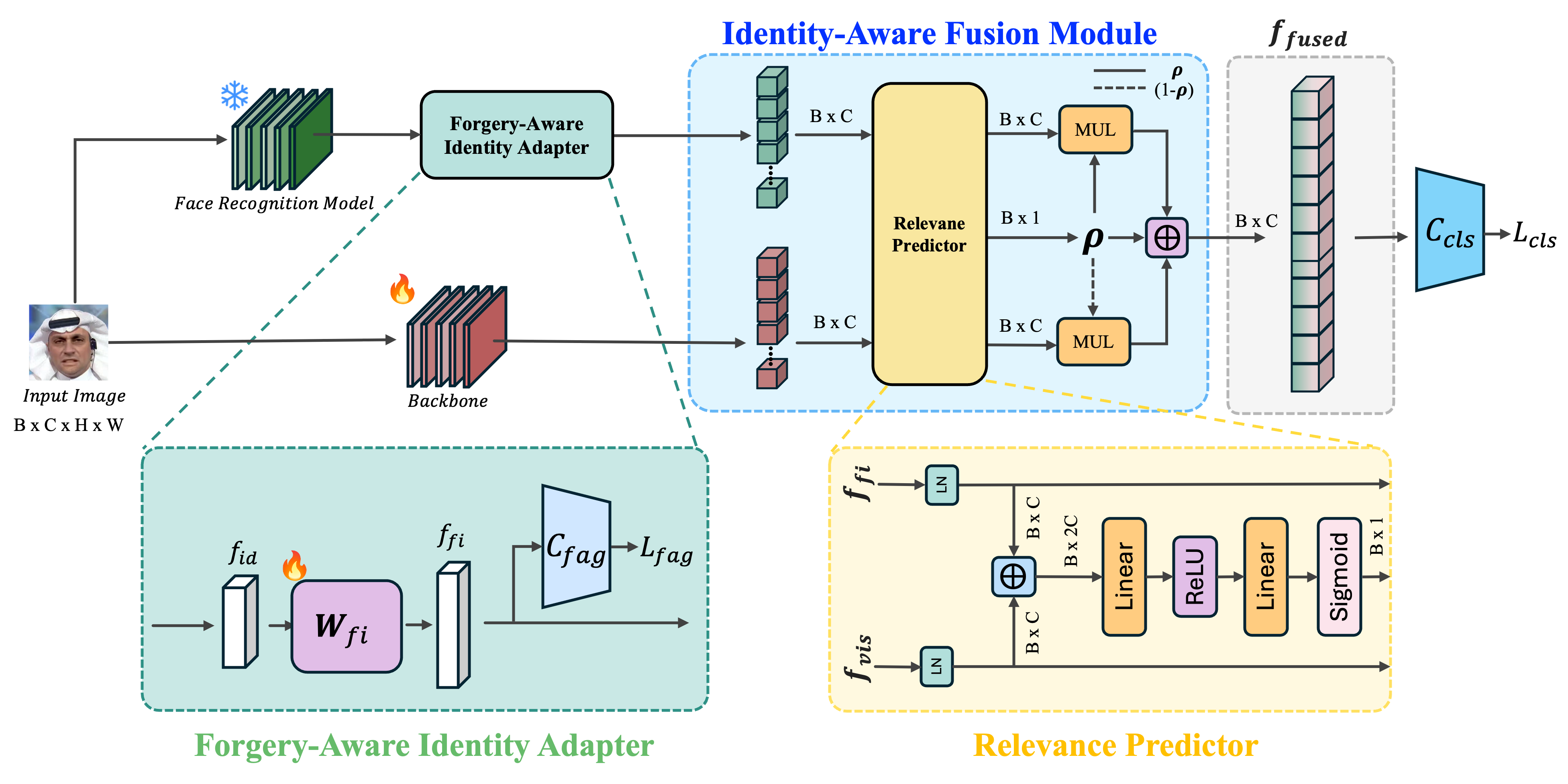}
    \caption{\textbf{Architecture of the proposed SELFI framework.} SELFI consists of two main modules: (1) the Forgery-Aware Identity Adapter (FAIA), which projects frozen identity embeddings into a forgery-relevant space using a trainable transformation and auxiliary supervision (Sec.~\ref{sec:faia}); and (2) the Identity-Aware Fusion Module (IAFM), which adaptively fuses identity and visual features based on a predicted relevance score (Sec.~\ref{sec:iafm}). By explicitly modeling and controlling identity usage, SELFI enables robust and generalizable deepfake detection across diverse manipulation types.}
    \label{fig:overview}
\end{figure*}

Motivated by the observations in Sec.~\ref{sec:preliminary}, we introduce \textbf{SELFI} (SELective Fusion of Identity), a framework designed for generalizable deepfake detection by adaptively fusing identity and visual cues. As illustrated in Fig.~\ref{fig:overview}, SELFI incorporates a \textbf{Forgery-Aware Identity Adapter (FAIA)} to transform identity embeddings into a forgery-relevant space and an \textbf{Identity-Aware Fusion Module (IAFM)} to dynamically integrate these with visual features based on estimated relevance. This adaptive fusion strategy allows SELFI to leverage reliable identity signals while suppressing biases, thereby enhancing generalization across diverse manipulation types, with detailed module designs presented in the following subsections.

\subsection{Forgery-Aware Identity Adapter (FAIA)}
\label{sec:faia}

The goal of FAIA is to extract face identity information from an input image and transform it into a representation that is more suitable for deepfake detection. While identity features from pretrained face recognition models are optimized for verifying who appears in an image, our preliminary analysis reveals that they can already serve as a surprisingly strong signal for deepfake detection. However, these embeddings are not specifically tailored for capturing forgery-related artifacts. To better align identity features with manipulation cues, FAIA learns to project them into a forgery-aware space where their utility for classification is maximized.

FAIA operates in two main stages. First, we extract a face identity embedding from the input image using a frozen face recognition model \( \mathcal{E}_{\text{id}} \):
\begin{equation}
\mathbf{f}_{\text{id}} = \mathcal{E}_{\text{id}}(\mathbf{x}),
\end{equation}
where \( \mathbf{x} \) is the input image and \( \mathbf{f}_{\text{id}} \in \mathbb{R}^{D_{\text{id}}} \) is the identity embedding. Here, \( D_{\text{id}} \) denotes the output dimensionality of the face recognition model. In our study, we use IResNet100~\cite{deng2019arcface}, for which \( D_{\text{id}} = 512 \).

Next, we project this identity embedding into the backbone feature space using a trainable weight matrix \( W_{\text{fi}} \in \mathbb{R}^{D_{\text{backbone}} \times D_{\text{id}}} \), where \( D_{\text{backbone}} \) denotes the feature dimensionality of the visual backbone. In our implementation, we use CLIP~\cite{radford2021learning} as the backbone, for which \( D_{\text{backbone}} = 768 \):
\begin{equation}
\mathbf{f}_{\text{fi}} = W_{\text{fi}} \mathbf{f}_{\text{id}}.
\end{equation}

To guide the transformation \( W_{\text{fi}} \) to produce identity features that are discriminative for forgery detection, we introduce an auxiliary supervision called the \textit{Forgery-Aware Guidance Loss}. Specifically, we attach a lightweight binary classifier \( \mathcal{C}_{\text{fag}} \) to the projected feature \( \mathbf{f}_{\text{fi}} \) and train it using the standard CrossEntropy loss:
\begin{equation}
\mathcal{L}_{\text{fag}} = \text{CE}(\mathcal{C}_{\text{fag}}(\mathbf{f}_{\text{fi}}), y),
\end{equation}
where \( y \in \{0,1\} \) indicates whether the input is real or fake.

This auxiliary supervision encourages \( W_{\text{fi}} \) to produce forgery-aware identity embeddings that are semantically aligned with manipulation cues. The resulting projected feature \( \mathbf{f}_{\text{fi}} \in \mathbb{R}^{D_{\text{backbone}}} \) is then fused with visual features in the Identity-Aware Fusion Module (IAFM).

\subsection{Identity-Aware Fusion Module (IAFM)}
\label{sec:iafm}

While forgery-aware identity features can be informative, they are not equally useful across all samples or manipulation types. Over-reliance on identity cues in irrelevant contexts may even hurt generalization. To address this, IAFM adaptively integrates identity and visual features by estimating the relevance of identity information for each input.

Given the projected identity feature \( \mathbf{f}_{\text{fi}} \in \mathbb{R}^{D_{\text{backbone}}} \) from FAIA and the visual feature \( \mathbf{f}_{\text{vis}} \in \mathbb{R}^{D_{\text{backbone}}} \) extracted from the visual encoder (CLIP~\cite{radford2021learning}), we first concatenate the two and feed them into a lightweight relevance predictor:
\begin{equation}
\rho = \mathcal{R}([\mathbf{f}_{\text{vis}} ; \mathbf{f}_{\text{fi}}]),
\end{equation}
where \( \mathcal{R}(\cdot) \) denotes a small feedforward network that outputs a scalar relevance score \( \rho \in [0, 1] \) via a sigmoid activation. In our implementation, \( \mathcal{R}(\cdot) \) consists of two linear layers with ReLU activation in between, followed by a sigmoid at the end:
\[
\mathcal{R}(\cdot) = \text{Sigmoid}(\text{Linear}_2(\text{ReLU}(\text{Linear}_1(\cdot)))).
\]

We then compute the final fused representation as a weighted combination of the identity and visual features:
\begin{equation}
\mathbf{f}_{\text{fused}} = \rho \cdot \mathbf{f}_{\text{fi}} + (1 - \rho) \cdot \mathbf{f}_{\text{vis}}.
\end{equation}

This soft fusion strategy enables the model to conditionally emphasize identity or visual cues depending on their relevance to the input, improving both flexibility and generalization. Finally, the fused feature \( \mathbf{f}_{\text{fused}} \) is passed to the final classification head \( \mathcal{C}_{\text{cls}} \) to produce the prediction:
\begin{equation}
\hat{y} = \mathcal{C}_{\text{cls}}(\mathbf{f}_{\text{fused}}),
\end{equation}
where \( \hat{y} \in [0, 1] \) represents the probability of the input being a fake.

\subsection{Overall Loss Function}
\label{sec:loss}

We jointly optimize two objectives: the main classification loss based on the fused representation and an auxiliary guidance loss applied to the projected identity features. The overall loss is defined as:
\begin{equation}
\mathcal{L}_{\text{total}} = \alpha \cdot \mathcal{L}_{\text{cls}} + \beta \cdot \mathcal{L}_{\text{fag}},
\end{equation}
where \( \mathcal{L}_{\text{cls}} \) is the standard CrossEntropy loss on the final prediction from \( \mathcal{C}_{\text{cls}} \), and \( \mathcal{L}_{\text{fag}} \) supervises the identity projection via \( \mathcal{C}_{\text{fag}} \). We use \( \alpha = 1.0 \) and \( \beta = 1.0 \) in all experiments.

%% file: main_tex/Experiments.tex
\section{Experiments}

\begin{table*}[t]
\centering
\caption{\textbf{Frame-level AUC Performance.} All models are trained on FF++ c23~\cite{rossler2019faceforensics++}. The best and second-best results are highlighted in \textbf{bold} and \underline{underlined}, respectively. Reported scores for prior methods are taken from DeepfakeBench~\cite{yan2023deepfakebench} and their original publications~\cite{cheng2024can, yan2024transcending}. A dash (–) indicates that the corresponding result is not available in the original papers.}

\label{tab:cross_dataset_frame_level}
\begin{adjustbox}{width=0.95\textwidth}
\begin{tabular}{l | c | c c c c c c}
\toprule
\textbf{Method} & \textbf{Publication} & \textbf{FF++ } & \textbf{CDFv2} & \textbf{DFD} & \textbf{DFDC} & \textbf{DFDCP} & \textbf{C-Avg.} \\
\midrule
Xception~\cite{chollet2017xception} & CVPR'17 & 0.964 & 0.737 & 0.816 & 0.708 & 0.737 & 0.750 \\
Meso4~\cite{afchar2018mesonet} & WIFS'18 & 0.608 & 0.609 & 0.548 & 0.556 & 0.599 & 0.578 \\
FWA~\cite{li2018exposing} & CVPRW'18 & 0.877 & 0.668 & 0.740 & 0.613 & 0.638 & 0.665 \\
EfficientB4~\cite{tan2019efficientnet} & ICML'19 & 0.957 & 0.749 & 0.815 & 0.696 & 0.728 & 0.747 \\
Capsule~\cite{nguyen2019capsule} & ICASSP'19 & 0.842 & 0.747 & 0.684 & 0.647 & 0.657 & 0.684 \\
CNN-Aug~\cite{wang2020cnn} & CVPR'20 & 0.849 & 0.702 & 0.646 & 0.636 & 0.617 & 0.650 \\
X-ray~\cite{li2020face} & CVPR'20 & 0.959 & 0.679 & 0.766 & 0.633 & 0.694 & 0.693 \\
FFD~\cite{dang2020detection} & CVPR'20 & 0.962 & 0.744 & 0.802 & 0.703 & 0.743 & 0.748 \\
F3Net~\cite{qian2020thinking} & ECCV'20 & 0.964 & 0.735 & 0.798 & 0.702 & 0.765 & 0.750 \\
SPSL~\cite{liu2021spatial} & CVPR'21 & 0.961 & 0.765 & 0.812 & 0.704 & 0.741 & 0.756 \\
SRM~\cite{luo2021generalizing} & CVPR'21 & 0.958 & 0.755 & 0.812 & 0.700 & 0.741 & 0.752 \\
CORE~\cite{ni2022core} & CVPRW'22 & 0.964 & 0.743 & 0.802 & 0.705 & 0.734 & 0.746 \\
Recce~\cite{cao2022end} & CVPR'22 & 0.962 & 0.732 & 0.812 & 0.713 & 0.742 & 0.750 \\
UCF~\cite{yan2023ucf} & ICCV'23 & \underline{0.971} & 0.753 & 0.807 & 0.719 & 0.759 & 0.760 \\
OPR~\cite{cheng2024can} & NeruIPS'24 & 0.959 & \textbf{0.845} & - & 0.724 & 0.812 & - \\
LSDA~\cite{yan2024transcending} & CVPR'24 & - & 0.830 & \underline{0.880} & \underline{0.736} & \underline{0.815} & \underline{0.815} \\
\midrule
SELFI (Ours) &  & \textbf{0.980} & \underline{0.839} & \textbf{0.907} & \textbf{0.796} & \textbf{0.840} & \textbf{0.846} \\
\bottomrule
\end{tabular}
\end{adjustbox}
\end{table*}

\subsection{Experimental Setting}

\textbf{Datasets.}
We train our models on the FaceForensics++ (FF++) dataset~\cite{rossler2019faceforensics++}, using the c23 (lightly compressed) version comprising four manipulation types: DeepFakes (DF)~\cite{faceswapdevs2019deepfakes}, Face2Face (F2F)~\cite{thies2016face2face}, FaceSwap (FS)~\cite{kowalski2018faceswap}, and NeuralTextures (NT). To evaluate cross-dataset generalization, we test on four disjoint datasets: Celeb-DF v2 (CDFv2)~\cite{li2020celeb}, DeepfakeDetection (DFD)~\cite{GoogleAI2019}, the Deepfake Detection Challenge (DFDC) and its preview version (DFDCP)~\cite{DeepfakeDetectionChallenge2025}. These benchmarks differ in content, manipulation methods, and post-processing, providing a comprehensive generalization testbed.

\textbf{Implementation Details.}
All experiments follow the training configuration and preprocessing procedures defined by the DeepfakeBench~\cite{yan2023deepfakebench} framework. We use CLIP as the backbone in our best-performing model. The input frames are resized to $224 \times 224$, and the model is trained with a batch size of 64 for 10 epochs. The hyperparameters $\alpha$ and $\beta$ are both set to 1. During training, we monitor the frame-level AUC on the validation set and save the model with the highest score. All experiments are conducted using five NVIDIA Titan RTX GPUs.


\subsection{Overall Performance on Comprehensive Datasets}
\textbf{Frame-level Generalization Performance.}
As shown in Tab.~\ref{tab:cross_dataset_frame_level}, we evaluate our model by training on FF++ and testing both in-domain (FF++) and cross-domain (CDFv2, DFD, DFDC, DFDCP) datasets. Our method achieves superior performance in the in-domain setting compared to all existing approaches. In the more challenging cross-domain scenario, our model outperforms all prior methods on every dataset except for CDFv2, where it ranks second. Notably, compared to LSDA~\cite{yan2024transcending}—one of the most recent and competitive state-of-the-art detectors—our approach achieves an average improvement of 3.5\% frame-level AUC across the cross-domain benchmarks, highlighting its strong generalization capability. A comparison with video-level deepfake detectors is provided in Sec.~\ref{appendix:video_level_auc}.

\begin{wraptable}{r}{0.5\linewidth}
\centering
\vspace{-2mm}
\caption{Cross-manipulation frame-level AUC results from~\cite{huang2023implicit}.}
\label{tab:cross_dataset_accuracy}
\begin{adjustbox}{width=0.5\textwidth}
\begin{tabular}{c|c|ccc|c}
\toprule
\textbf{Train} & \textbf{Method} & \textbf{DF} & \textbf{FS} & \textbf{FST} & \textbf{Avg} \\
\midrule
\multirow{6}{*}{DF}
 & MAT~\cite{zhao2021multi}   & \underline{0.999} & 0.406 & 0.454 & 0.620 \\
 & GFF~\cite{luo2021generalizing}   & \underline{0.999} & 0.472 & 0.519 & 0.663 \\
 & DCL~\cite{sun2022dual}   & \textbf{1.000} & 0.610 & \underline{0.685} & 0.765 \\
 & IID~\cite{huang2023implicit}   & 0.995 & 0.638 & \textbf{0.735} & \underline{0.789} \\
 & CLIP~\cite{radford2021learning}  & \underline{0.999} & \underline{0.731} & 0.625 & 0.785 \\
 & Ours  & \underline{0.999} & \textbf{0.809} & 0.657 & \textbf{0.822} \\
\midrule
\multirow{6}{*}{FS}
 & MAT~\cite{zhao2021multi}   & 0.641 & \underline{0.997} & 0.574 & 0.737 \\
 & GFF~\cite{luo2021generalizing}   & 0.702 & \underline{0.997} & 0.613 & 0.771 \\
 & DCL~\cite{sun2022dual}   & 0.748 & \textbf{0.999} & 0.649 & 0.799 \\
 & IID~\cite{huang2023implicit}   & 0.754 & \underline{0.997} & \underline{0.662} & 0.804 \\
 & CLIP~\cite{radford2021learning}  & \underline{0.906} & 0.995 & 0.642 & \underline{0.848} \\
 & Ours  & \textbf{0.935} & 0.995 & \textbf{0.733} & \textbf{0.888} \\
\midrule
\multirow{6}{*}{FST}
 & MAT~\cite{zhao2021multi}   & 0.582 & 0.550 & 0.992 & 0.708 \\
 & GFF~\cite{luo2021generalizing}   & 0.615 & 0.562 & 0.915 & 0.724 \\
 & DCL~\cite{sun2022dual}   & 0.640 & 0.584 & \underline{0.995} & 0.740 \\
 & IID~\cite{huang2023implicit}   & 0.654 & \underline{0.595} & \underline{0.995} & \underline{0.748} \\
 & CLIP~\cite{radford2021learning}  & \underline{0.724} & 0.521 & \textbf{0.996} & 0.747 \\
 & Ours  & \textbf{0.803} & \textbf{0.626} & 0.992 & \textbf{0.807} \\
\bottomrule
\end{tabular}
\end{adjustbox}
\end{wraptable}

\textbf{Cross-Manipulation Performance.}
To assess the generalization ability of SELFI, we evaluate its performance in a cross-manipulation setting where the model is trained on one manipulation type and tested on others. As shown in Tab.~\ref{tab:cross_dataset_accuracy}, SELFI consistently achieves the highest average AUC across all training settings. When trained on DeepFakes, SELFI improves the average AUC by 3.7\% over CLIP and 3.3\% over IID. With FaceSwap as the training source, SELFI achieves the best overall performance, surpassing CLIP by 4.0\% in average AUC. Even when trained on FaceShifter, which is generally more challenging for generalization, SELFI still outperforms all competitors, recording a 2.8\% gain over the best baseline. These results demonstrate that naive identity injection or static fusion strategies are insufficient for robust cross-manipulation detection. In contrast, SELFI's adaptive integration of identity information via a learned relevance score allows the model to effectively suppress manipulation-specific biases, leading to superior generalization on unseen forgeries.

\begin{table}[t]
\centering
\vspace{-2mm}
\begin{minipage}[t]{0.48\linewidth}
\centering
\caption{\textbf{Ablation study of SELFI modules with CLIP backbone.} FAIA-only denotes simple feature concatenation without adaptive fusion. (see Sec.~\ref{appendix:module_ablation} for details)}

\label{tab:selfi_ablation_clip_left}
\begin{adjustbox}{width=\textwidth}
\begin{tabular}{ccc|cccc}
\toprule
\multicolumn{3}{c|}{\textbf{Modules}} & \multicolumn{4}{c}{\textbf{CLIP}} \\
\cmidrule(lr){1-3} \cmidrule(lr){4-7}
FAIA & $\mathcal{L}_{\text{fag}}$ & IAFM & CDFv2 & DFD & DFDCP & Avg \\
\midrule
\ding{55} & \ding{55} & \ding{55} & 0.795 & 0.830 & 0.793 & 0.806 \\
\ding{51} & \ding{55} & \ding{55} & 0.804 & 0.887 & \textbf{0.844} & 0.845 \\
\ding{51} & \ding{55} & \ding{51} & \underline{0.819} & \underline{0.899} & 0.830 & \underline{0.849} \\
\ding{51} & \ding{51} & \ding{51} & \textbf{0.839} & \textbf{0.907} & \underline{0.840} & \textbf{0.862} \\
\bottomrule
\end{tabular}
\end{adjustbox}
\end{minipage}
\hfill
\begin{minipage}[t]{0.48\linewidth}
\centering
\caption{\textbf{Cross-dataset AUC of SELFI with different backbones.} SELFI consistently improves generalization across all architectures.}
\label{tab:selfi_ablation_clip_right}
\begin{adjustbox}{width=0.95\textwidth}
\begin{tabular}{c|cccc}
\toprule
\textbf{Backbones} & \textbf{CDFv2} & \textbf{DFD} & \textbf{DFDCP} & \textbf{Avg} \\
\midrule
CLIP & 0.795 & 0.830 & 0.793 & 0.806 \\
+SELFI  & \textbf{0.839} & \textbf{0.907} & \textbf{0.840} & \textbf{0.862} \\
\midrule
ResNet34 & 0.739 & 0.810 & 0.690 & 0.746 \\
+SELFI  & \textbf{0.762} & \textbf{0.822} & \textbf{0.731} & \textbf{0.772} \\
\midrule
EfficientB4 & 0.747 & \textbf{0.823} & 0.685 & 0.751 \\
+SELFI  & \textbf{0.755} & 0.812 & \textbf{0.733} & \textbf{0.767} \\
\bottomrule
\end{tabular}
\end{adjustbox}
\end{minipage}
\end{table}

\subsection{Ablation Study}
\textbf{Module-Wise Contribution Analysis.}
To verify the contribution of each component within SELFI, we perform a step-by-step ablation using CLIP as the backbone (Tab.~\ref{tab:selfi_ablation_clip_left}). Incorporating only the Forgery-Aware Identity Adapter (FAIA), where identity features are projected and simply concatenated with backbone features before classification, improves average AUC by 3.9\% over the baseline. This suggests that even unmodulated identity features can serve as useful cues when aligned with visual content. The detailed architecture of this configuration is described in Sec.~\ref{appendix:module_ablation}. Introducing the Identity-Aware Fusion Module (IAFM), which adaptively weighs the relevance of identity information, further improves performance by 0.4\%, highlighting the benefit of conditional identity integration. Finally, when supervised with the forgery-specific guidance loss $\mathcal{L}_{\text{fs}}$, the full SELFI framework achieves an additional 1.3\% gain, validating the synergistic effect of jointly learning forgery-aware identity representations and adaptive fusion. For additional ablation results on other backbones, please refer to Sec.~\ref{appendix:module_ablation}.

\textbf{Effect of Auxiliary Feature Source.}
To assess whether the observed performance gains stem from meaningful identity information or are simply due to feature ensemble effects, we conduct an ablation study using various auxiliary sources. As shown in Tab.~\ref{tab:source_ablation}, we fix the deepfake detection backbone to CLIP and vary the auxiliary source while keeping the feature extractor architecture consistent (see Sec.~\ref{appendix:source_ablation} for details). We consider four types of auxiliary sources: (1) Random Initialization, representing purely uninformative features; (2) ImageNet, trained on generic object recognition and not specific to faces; (3) Deepfake-Trained, trained to detect forgeries but lacking explicit identity information; and (4) Face Identity, our proposed identity feature extracted from a frozen face recognition model (IResNet100). The results clearly show that identity-aware features lead to the best generalization, boosting performance by 23.8\% over Random, 22.9\% over ImageNet, and 13.9\% over Deepfake-Trained features. These findings validate that SELFI’s fusion benefits arise not from ensemble size, but from the quality and relevance of the auxiliary signal—highlighting that identity semantics, not just forgery correlations, are essential for robust and transferable deepfake detection.


\textbf{Generalization Across Backbones.}
To evaluate the versatility of SELFI, we integrate it with three different backbone architectures—CLIP, ResNet34, and EfficientNet-B4—and compare their performance with and without SELFI (Tab.~\ref{tab:selfi_ablation_clip_right}). Across all configurations, SELFI consistently improves cross-dataset AUC: CLIP shows the largest gain with a 5.6\% improvement, followed by ResNet34 with 2.6\% and EfficientNet-B4 with 1.6\%. These results demonstrate that SELFI is highly transferable and can be seamlessly plugged into diverse feature extractors, offering consistent improvements regardless of model capacity or architecture. This backbone-agnostic behavior underscores the practical utility of SELFI as a general-purpose deepfake detection enhancement module.

\begin{table*}[t]
\centering
\caption{\textbf{Effect of auxiliary feature sources on frame-level AUC.} Only identity-aware features improve generalization, while others offer limited benefit.}

\label{tab:source_ablation}
\begin{adjustbox}{width=0.85\textwidth}
\begin{tabular}{l | c c c c c c}
\toprule
\textbf{Auxiliary Source} & \textbf{FF++ } & \textbf{CDFv2} & \textbf{DFD} & \textbf{DFDC} & \textbf{DFDCP} & \textbf{C-Avg.} \\
\midrule
X (No Fusion) & \underline{0.979} & \underline{0.795}  & \underline{0.830} & \underline{0.810} & \underline{0.793} & \underline{0.807} \\
Random Intialization & 0.542  & 0.544 & 0.513 & 0.552 & 0.581 & 0.548 \\
ImageNet & 0.726  & 0.586 & 0.650 & 0.617 & 0.615 & 0.617 \\
Deepfake-Trained & 0.975  & 0.697 & 0.771 & 0.693 & 0.666 & 0.707\\
Face Identity & \textbf{0.980} & \textbf{0.839}  & \textbf{0.907} & \textbf{0.796} & \textbf{0.840} & \textbf{0.846} \\
\bottomrule
\end{tabular}
\end{adjustbox}
\end{table*}


\textbf{Limitations.}
Although SELFI adaptively fuses identity and visual features based on a learned relevance score, its effectiveness may be influenced by the quality of the extracted identity embeddings. In cases where the face is partially visible or captured from extreme angles (e.g., side profiles), identity representations may be less reliable, potentially affecting relevance estimation. Furthermore, while our work focuses on identity-related bias as a primary factor, other sources of bias—such as background, ethnicity, or compression artifacts—have also been discussed in prior studies~\cite{yan2023ucf, lin2024preserving}. A comprehensive analysis of these factors is beyond the scope of this work, but presents a valuable direction for future exploration.

\section{Conclusion}
This work presents SELFI, a novel framework for generalizable deepfake detection that explicitly and adaptively incorporates face identity information. Through extensive analysis, we show that identity features—when used indiscriminately—can act as both powerful cues and harmful biases, depending on the manipulation context. Motivated by this observation, SELFI separates identity cues from visual content and learns to fuse them based on per-sample relevance, leveraging their benefits while mitigating overfitting. Our method consistently outperforms strong baselines across both in-domain and cross-domain settings, and demonstrates robust performance across diverse backbones. Furthermore, ablation studies confirm that SELFI’s gains stem from semantically meaningful identity information, rather than naive feature ensembling. We believe these insights offer a principled and practical direction for enhancing the generalization ability of deepfake detectors, and encourage future work to extend adaptive feature integration to other sources of bias beyond identity.

%% file: main_tex/appendix.tex
\appendix
\section*{\centering\LARGE Supplementary Material}

\section{Appendix}
\subsection{Details of Experimental Setting for Sec.~\ref{sec:preliminary}}
\label{appendix:preliminary}

\begin{figure*}[h]
\centering
    \includegraphics[width=0.6\textwidth]{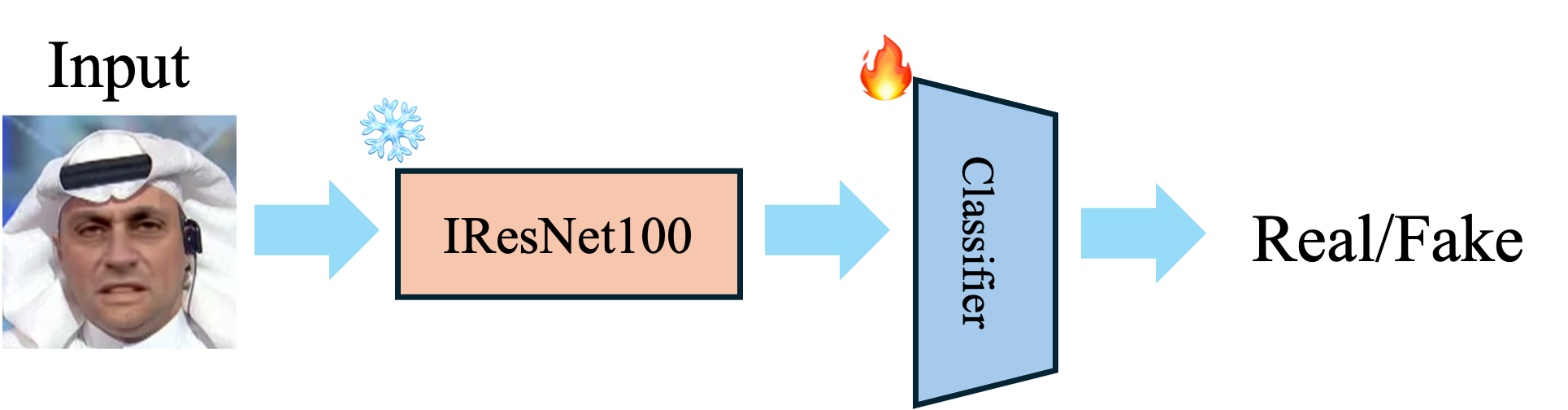}
    \caption{
    \textbf{Architecture of the identity-only detection setting.} A lightweight classifier is trained on top of frozen face identity embeddings extracted from a face recognition model, without any access to visual content. This setup is used to investigate whether face identity alone can support deepfake detection.
    }
\label{fig:identity_only_architecture}
\end{figure*}

To investigate whether face identity alone can support deepfake detection, we design an identity-only detection setup where no visual content is used. Specifically, we extract identity embeddings from a pretrained face recognition model (IResNet100), which is frozen throughout training. A lightweight classifier head is then trained on top of these embeddings to classify whether the input is real or fake. The overall architecture is illustrated in Fig.~\ref{fig:identity_only_architecture}. Since the output of IResNet100 is a 512-dimensional feature vector, we append a linear classifier that maps this 512-dimensional identity representation into a 2-class prediction space (real or fake).

\begin{table*}[h]
\centering
\caption{
\textbf{Cross-manipulation performance of identity-only detection.} Each model is trained using identity features extracted from a frozen face recognition model, without visual content, and evaluated across different manipulation types. Diagonal entries represent in-domain performance, while off-diagonal entries indicate cross-manipulation generalization. Results show that some manipulations (e.g., DF, FS) retain transferable identity cues, whereas others (e.g., NT) lead to poor generalization.
}\label{tab:appendix_source_ablation}
\begin{adjustbox}{width=0.5\textwidth}
\begin{tabular}{l | c c c c |c}
\toprule
\textbf{Train} & \textbf{NT} & \textbf{FS} & \textbf{F2F} & \textbf{DF} & \textbf{Avg} \\
\midrule
NT & \textbf{0.608}  & 0.535 & \underline{0.569} & 0.540 & 0.563 \\
FS & 0.509  & \textbf{0.846} & 0.565 & \underline{0.745} & \underline{0.666} \\
F2F & \underline{0.524}  & 0.509 & \textbf{0.826} & 0.523 & 0.595\\
DF & {0.519} & \underline{0.721}  & {0.537} & \textbf{0.901} & \textbf{0.670} \\
\midrule
All & 0.550 & 0.756 & 0.714 & 0.807 & {0.707} \\
\bottomrule
\end{tabular}
\end{adjustbox}
\end{table*}

We conduct cross-manipulation experiments using the FaceForensics++ dataset, focusing on four manipulation types: Deepfake (DF), FaceSwap (FS), Face2Face (F2F), and NeuralTextures (NT). In each setting, we train the classifier using identity features from one manipulation method and evaluate its generalization to the remaining ones. This setup allows us to examine how well identity-based signals transfer across manipulation types. The full results are reported in Tab.~\ref{tab:appendix_source_ablation}.

For all experiments, we use the standard CrossEntropy loss and train the classifier with the AdamW optimizer. The learning rate is set to 0.0002. Each model is trained for 10 epochs, and we save the best model based on frame-level AUC on the validation set.

\subsection{Video-level AUC Comparisoon}
\label{appendix:video_level_auc}

\begin{table}[h]
\centering
\caption{\textbf{Video-level AUC Performance.} Scores for other methods are reported from~\cite{haliassos2022leveraging} and their respective original publications~\cite{zhao2023istvt, wang2023noise, zhang2024learning}.}
\label{tab:cross_dataset_video_level}
\begin{adjustbox}{width=0.65\textwidth}  
\begin{tabular}{l | c | ccc}
  \toprule
  \textbf{Method} & \textbf{Publication} & \textbf{CDFv2} & \textbf{DFDC} & \textbf{Avg} \\
  \midrule
  LipForensics~\cite{haliassos2021lips} & CVPR'21 & 0.824 & 0.735 & 0.780 \\
  FTCN~\cite{zheng2021exploring} & ICCV'21 & 0.869 & 0.740 & 0.805 \\
  RealForensics~\cite{haliassos2022leveraging} & CVPR'22 & 0.869 & 0.759 & 0.814 \\
  ISTVT~\cite{zhao2023istvt} & TIFS'23 & 0.841 & 0.742 & 0.792 \\
  NoiseDF~\cite{wang2023noise} & AAAI'23 & 0.759 & 0.639 & 0.699 \\
  NACO~\cite{zhang2024learning} & ECCV'24 & \textbf{0.895} & \underline{0.767} & \underline{0.831} \\
  \midrule
  SELFI (Ours) &  & \underline{0.893} & \textbf{0.820} & \textbf{0.857} \\
  \bottomrule
\end{tabular}
\end{adjustbox}
\end{table}

In addition to frame-level evaluations, we compare our method against recent state-of-the-art video-level detectors, as summarized in Tab.~\ref{tab:cross_dataset_video_level}. Since our model operates at the frame level, we follow the evaluation protocol commonly used in other frame-level studies, where 32 evenly sampled frames are extracted from each video and individually classified. The final video-level prediction is then obtained by averaging the frame-level outputs. While our method slightly underperforms the strongest baseline on CDFv2, it achieves the best performance on DFDC and records the highest average video-level AUC overall. Notably, our method surpasses NACO~\cite{zhang2024learning}—the most recent and competitive video-level detector—by 2.6\% in average AUC, despite relying solely on frame-level information without any explicit modeling of temporal dynamics. This demonstrates the strength of our spatial feature representation in generalizing across videos without temporal supervision.

\subsection{Details of Experimental Setting for Tab.~\ref{tab:selfi_ablation_clip_left}}
\label{appendix:module_ablation}

\begin{table}[h]
\centering
\caption{Ablation study of SELFI framework on EfficientNetb4 and ResNet34. We incrementally add FAIA, $\mathcal{L}_{\text{fs}}$, and IAFM to examine their contributions. Best results per backbone are \textbf{bolded}.}
\resizebox{\textwidth}{!}{
\begin{tabular}{ccc|cccc|cccc}
\toprule
\multicolumn{3}{c|}{\textbf{Modules}} & \multicolumn{4}{c|}{\textbf{EfficientNetb4}} & \multicolumn{4}{c}{\textbf{ResNet34}} \\
\cmidrule(lr){1-3} \cmidrule(lr){4-7} \cmidrule(lr){8-11}
FAIA & $\mathcal{L}_{\text{fs}}$ & IAFM & CDFv2 & DFD & DFDCP & Avg & CDFv2 & DFD & DFDCP & Avg \\
\midrule
\ding{55} & \ding{55} & \ding{55} & \underline{0.747} & \underline{0.823} & 0.685 & 0.751 & 0.739 & 0.810 & 0.690 & 0.746 \\
\ding{51} & \ding{55} & \ding{55} & 0.737 & \textbf{0.851} & 0.689 & \underline{0.759} & \underline{0.755} & 0.813 & 0.678 & 0.749 \\
\ding{51} & \ding{55} & \ding{51} & 0.721 & 0.818 & \underline{0.695} & 0.745 & 0.739 & \underline{0.814} & \underline{0.730} & \underline{0.761} \\
\ding{51} & \ding{51} & \ding{51} & \textbf{0.755} & 0.812 & \textbf{0.733} & \textbf{0.767} & \textbf{0.762} & \textbf{0.822} & \textbf{0.731} & \textbf{0.772} \\
\bottomrule
\end{tabular}
}
\label{fig:module_backbone_ablation}

\end{table}

To analyze the contribution of each component within the SELFI framework, we perform a step-by-step ablation study under four configurations. The first setting disables all modules (FAIA, $\mathcal{L}_{\text{fs}}$, and IAFM), where the model is trained solely using the backbone feature. The second setting enables only FAIA, where identity embeddings are projected and simply concatenated with visual features before classification—without using the relevance predictor or soft fusion. In the third setting, both FAIA and IAFM are activated, allowing conditional fusion via the relevance predictor, but without the forgery-aware guidance loss. The final setting corresponds to the full SELFI model, which combines all modules.

In addition to CLIP, we also evaluate the ablations on two alternative backbones—EfficientNet-B4 and ResNet34—to assess the consistency of SELFI's modular benefits across architectures. The results are presented in Tab.~\ref{fig:module_backbone_ablation}, showing that each module incrementally contributes to performance improvements, and that the full SELFI configuration achieves the highest average AUC on both backbones. Notably, adding only FAIA yields a noticeable gain over the baseline, demonstrating that identity embeddings alone—when properly aligned—are useful for detection. Incorporating IAFM further improves performance by adaptively weighing identity relevance. Finally, the inclusion of the forgery-aware guidance loss $\mathcal{L}_{\text{fs}}$ consistently enhances results, confirming the importance of supervising identity transformation with task-specific signals. These trends are consistent across both EfficientNet-B4 and ResNet34, validating SELFI’s backbone-agnostic effectiveness.

For both EfficientNet-B4 and ResNet34, we use input resolution of 256×256 and train models using the Adam optimizer with a learning rate of 0.0002, $\beta_1 = 0.9$, $\beta_2 = 0.999$, and weight decay of 0.0005. All models are trained for 10 epochs, and the best model is selected based on validation AUC. Both backbones are initialized with ImageNet pretraining.

\subsection{Details of Experimental Setting for Tab.~\ref{tab:source_ablation}}
\label{appendix:source_ablation}

\begin{table*}[h]
\centering
\caption{Standalone performance of the Deepfake-Trained ResNet101 model used as an auxiliary source in Tab.~\ref{tab:source_ablation}. All results are reported as frame-level AUC.}
\label{tab:source_ablation_appendix}
\begin{adjustbox}{width=0.8\textwidth}
\begin{tabular}{l | c c c c c c}
\toprule
\textbf{Method} & \textbf{FF++ } & \textbf{CDFv2} & \textbf{DFD} & \textbf{DFDC} & \textbf{DFDCP} & \textbf{C-Avg.} \\
\midrule
ResNet101 & {0.977} & {0.720}  & {0.781} & {0.693} & {0.680} & {0.719} \\
\bottomrule
\end{tabular}
\end{adjustbox}
\end{table*}

All auxiliary sources used in Tab.~\ref{tab:source_ablation} are extracted from the same backbone architecture, ResNet101, to ensure fair comparison. For the \textbf{Face Identity} extractor, we utilize a pretrained IResNet100 model, which applies Batch Normalization to its output features. To maintain consistency, we apply Batch Normalization to the output features of all other ResNet101-based extractors before feeding them into the IAFM module.

The \textbf{Deepfake-Trained} ResNet100 is trained separately using a standard deepfake classification setup. Its standalone performance (i.e., without fusion) is reported in Tab.~\ref{tab:source_ablation_appendix} for reference. All auxiliary feature extractors are frozen during SELFI training and only their projected features are used for fusion with CLIP features.